\documentclass[10pt,twocolumn,letterpaper]{article}

\usepackage{subfig}

\usepackage{cvpr}
\usepackage{times}
\usepackage{epsfig}
\usepackage{graphicx}
\usepackage{amsmath}
\usepackage{amssymb}
\usepackage{multirow}
\usepackage{tabu}
\usepackage{booktabs}
\usepackage{enumitem}

\cvprfinalcopy 

\def\cvprPaperID{2938} 

\ifcvprfinal\fi
\begin{document}

\title{GroupFace: Learning Latent Groups and Constructing Group-based Representations for Face Recognition}

\author{Yonghyun Kim$^1$ \quad Wonpyo Park$^2$ \quad Myung-Cheol Roh$^1$ \quad Jongju Shin$^1$
\smallskip
\\
$^1$Kakao Enterprise, $^2$Kakao Corp.
}

\maketitle

\begin{abstract}
    In the field of face recognition, a model learns to distinguish millions of face images with fewer dimensional embedding features, and such vast information may not be properly encoded in the conventional model with a single branch.
    We propose a novel face-recognition-specialized architecture called GroupFace that utilizes multiple group-aware representations, simultaneously, to improve the quality of the embedding feature.
    The proposed method provides self-distributed labels that balance the number of samples belonging to each group without additional human annotations, and learns the group-aware representations that can narrow down the search space of the target identity.
    We prove the effectiveness of the proposed method by showing extensive ablation studies and visualizations.
    All the components of the proposed method can be trained in an end-to-end manner with a marginal increase of computational complexity.
    Finally, the proposed method achieves the state-of-the-art results with significant improvements in 1:1 face verification and 1:N face identification tasks on the following public datasets: LFW, YTF, CALFW, CPLFW, CFP, AgeDB-30, MegaFace, IJB-B and IJB-C.
\end{abstract}

\section{Introduction}

Face recognition is a primary technique in computer vision to model and understand the real world.
Many methods and enormous datasets \cite {vggface2, msceleb1m, kemelmacher2016megaface, vggface, imdbface, casia_webface} have been introduced, and recently, methods that use deep learning \cite{ArcFace, uniformface, AFRN, SphereFace, CosFace, centerloss, regularface} have greatly improved the face recognition accuracy, but it still falls short of expectations.

To reduce the shortfall, most of the recent research in face recognition focused on improving the loss function.
The streams from CenterLoss \cite{centerloss}, CosFace \cite{CosFace}, ArcFace \cite{ArcFace} and RegularFace \cite{regularface} all tried to minimize the intra-class variation and maximize the inter-class variation.
These methods are effective and have gradually improved the accuracy by elaborating the objective of learning.

Despite the development of loss functions, general-purpose networks, not a network devised for a face recognition, can have difficulty in enabling effective training of the network to recognize a huge number of person identities.
Unlike common problems such as classification, in the evaluation stage, a face-recognition model encounters new identities, which are not included in the training set.
Thus, the model has to embed nearly 100k identities \cite{msceleb1m} in the training set and also consider a huge number of unknown identities.
However, most of the existing methods just attach several fully-connected layers after widely-used backbone networks such as VGG \cite{vggface} and ResNet \cite{resnet} without any designs for the characteristics of face recognition.

\begin{figure}[t]
    \begin{center}
        \includegraphics[width=8.4cm]{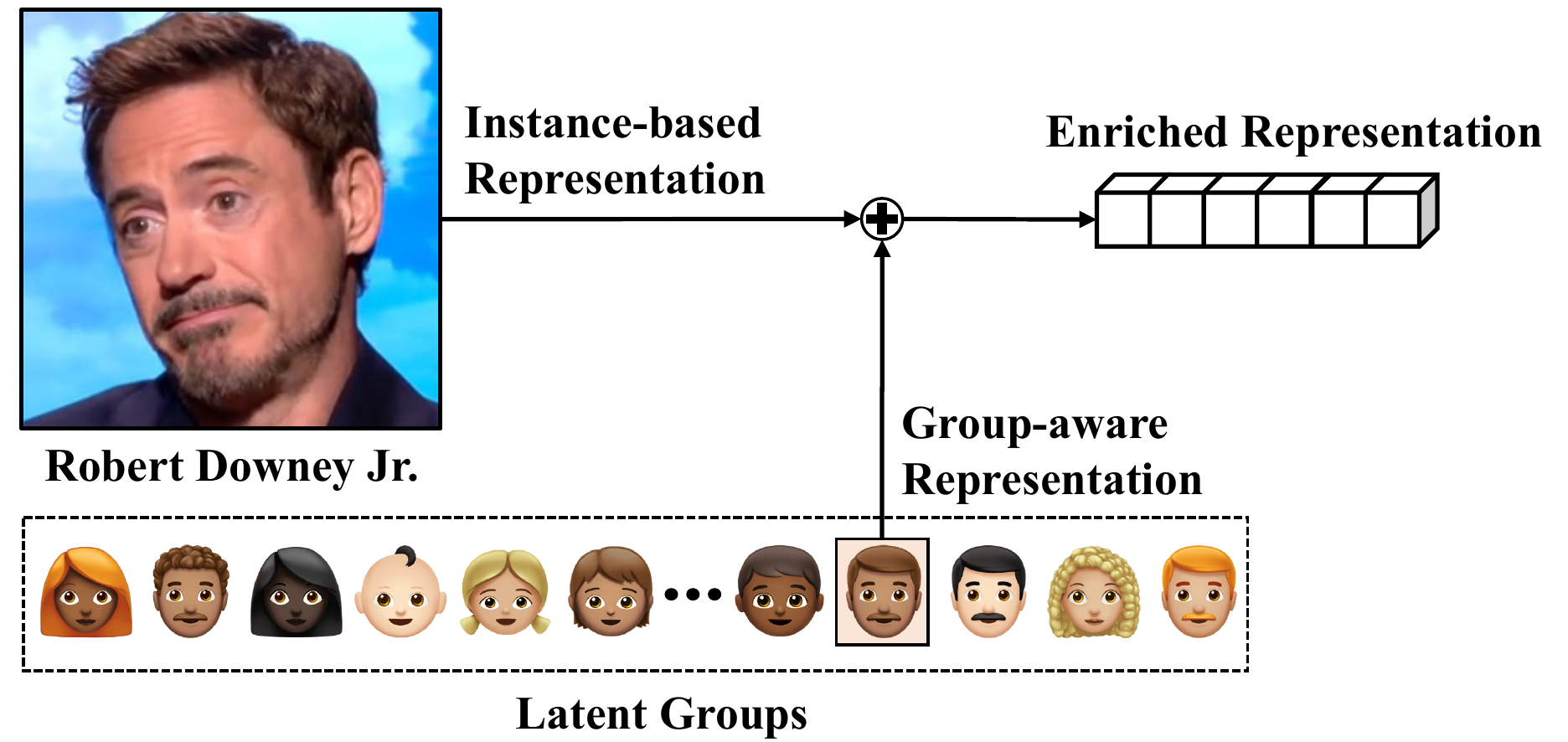}
    \end{center}
    \caption{
        Conceptual scheme of the proposed method.
        The proposed method enhances the quality of the embedding feature by supporting the instance-based representation of \textbf{Robert} \textbf{Downey} \textbf{Jr.} using the group-aware representation considering \textbf{Brown-beard} \textbf{Man} \textbf{Group}.
        }
    \label{fig:concept}
\end{figure}

Grouping is a key idea to efficiently-and-flexibly embed a significant number of people and briefly describe an unknown person.
Each person has own characteristics in his or her face.
At the same time, they have common ones shared in a group of people.
In the real world, group-based description (man with deep, black eyes and red beard) that involves common characteristics in the group, can be useful to narrow down the set of candidates, even though it cannot identify the exact person.
Unfortunately, explicit grouping requires manual categorizing on huge data and may be limited by the finite range of descriptions by human knowledge,
However, by adopting the concept of grouping, the recognition network can reduce the search space and flexibly embed a significant number of identities into an embedding feature.

We propose a novel face-recognition architecture called GroupFace that learns multiple latent groups and constructs group-aware representations to effectively adopt the concept of grouping (Figure \ref{fig:concept}).
We define Latent Groups, which are internally determined as latent variables by comprehensively considering facial factors (\eg, hair, pose, beard) and non-facial factors (\eg, noise, background, illumination).
To learn the latent groups, we introduce a self-distributed grouping method that determines group labels by considering the overall distribution of latent groups.
The proposed GroupFace structurally ensembles multiple group-aware representations into the original instance-based representation for face recognition.

We summarize the contributions as follows:
\begin{itemize}[leftmargin=+.2in,label=$\bullet$]
    \item
    GroupFace is a novel face-recognition-specialized architecture that integrates the group-aware representations into the embedding feature and provides well-distributed group-labels to improve the quality of feature representation.
    GroupFace also suggests a new similarity metric to consider the group information additionally.
    \item
    We prove the effectiveness of GroupFace in extensive experiments and ablation studies on the behaviors of GroupFace. 
    \item
    GroupFace can be applied many existing face-recognition methods to obtain a significant improvement with a marginal increase in the resources.
    Especially, a hard-ensemble version of GroupFace can achieve high recognition-accuracy by adaptively using only a few additional convolutions. 
\end{itemize}




\section {Related Works}

\noindent\textbf{Face Recognition} has been studied for decades.
Many researchers proposed machine learning techniques with feature engineering \cite{10.1007/978-3-540-24670-1_36, 6619233, joint_bayesian, fisherface, 5459250,CSML, eigenface, WHT:ECCVW08:DBMW,QiYin:2011:AMF:2191740.2192084}.
Recently, deep learning methods have overcome the limitations of traditional face-recognition approaches with public face-recognition datasets \cite{vggface2, msceleb1m, kemelmacher2016megaface, vggface, imdbface, casia_webface}.
DeepFace \cite{deepface} used 3D face frontalization to achieve a breakthrough in face recognition methods that use deep learning.
FaceNet \cite{FaceNet} proposed triplet loss to maximize the distance between an anchor and its negative sample, and to minimize the distance between the same anchor and its positive sample.
CenterLoss \cite{centerloss} proposed center loss to minimize the distance between samples and their class centers.
MarginalLoss \cite{marginal_loss} adopted the concept of margin to minimize intra-class variations and to keep inter-class distances with margin.
RangeLoss \cite{range_loss} used long-tailed data during the training stage.
RingLoss \cite{ring_loss} constrained a feature's magnitude to be a certain number.
NormFace \cite{NormFace} proposed to normalize features and fully connected layer weights; verification accuracy increased after normalization.
SphereFace \cite{SphereFace} proposed angular softmax (A-Softmax) loss with multiplicative angular margin.
Based on A-Softmax, CosFace \cite{CosFace} proposed an additive cosine margin and ArcFace \cite{ArcFace} applies an additive angular margin.
The authors of RegularFace \cite{regularface} and UniformFace \cite{uniformface} argued that approaches that use angular margin \cite{ArcFace, SphereFace, CosFace} concentrated on intra-class compactness only, then suggested new losses to increase the inter-class variation.
These previous methods, in general, focused on how to improve loss functions to improve face recognition accuracy with conventional feature representation.
A slight change such as adding a few layers or increasing the number of channels, commonly did not bring a noticeable improvement.
However, GroupFace improves the quality of feature representation and achieves a significant improvement by adding a few more layers in parallel.
\smallbreak

\noindent\textbf{Grouping} or clustering methods such as k-means internally categorize samples by considering relative metrics such as a cosine similarity or Euclidean distance without explicit class labels.
In general, these clustering methods attempt to to construct well-distinguished categories by preventing the assignment of most images to one or a few clusters. Recently, several methods that used deep learning \cite{caron2018deep,noroozi2016unsupervised,yang2016joint} have been introduced.
These methods are effective, however, they use full batches as in previous methods, not mini-batches as in deep learning.
Thus, these methods are not readily incorporate deeply and end-to-end in an application framework. 
To efficiently learn the latent groups, our method introduces a self-distributed grouping method that considers an expectation-normalized probability in a deep manner.


\begin{figure*}[ht!]
    \begin{center}
    \includegraphics[width=17cm]{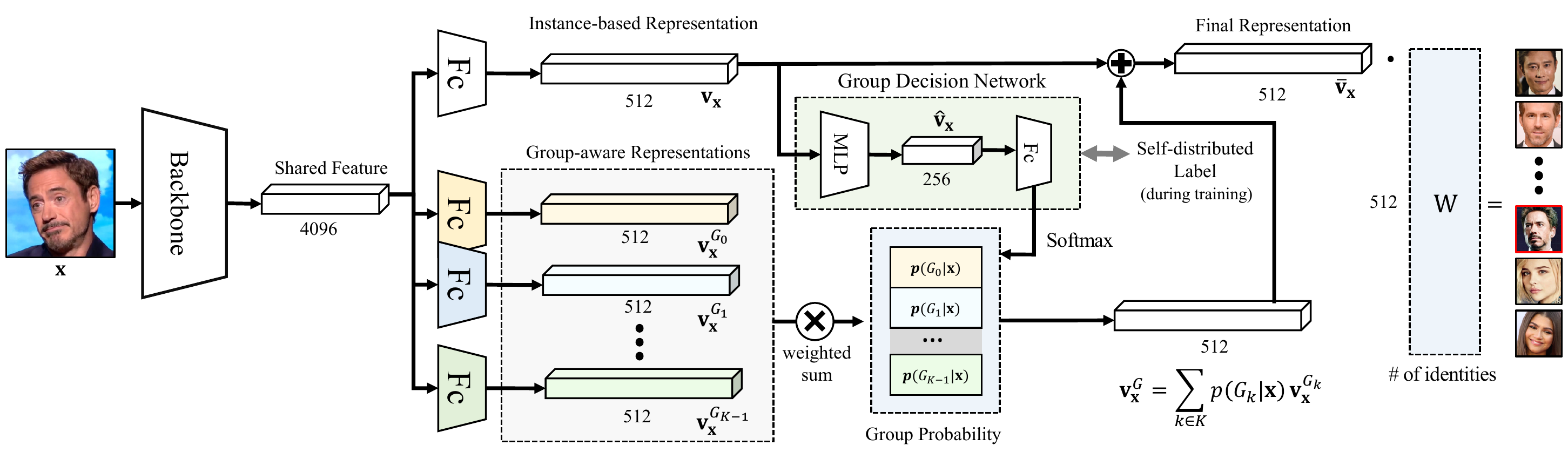}
    \end{center}
       \caption{GroupFace generates a shared feature of 4096 dimension and deploys a fully-connected layer for an instance-based representation $\mathbf{v}_{\mathbf{x}}$ and $K$ fully-connected layers for group-aware representations $\mathbf{v}_{\mathbf{x}}^{{G}}$ for a given sample $\mathbf{x}$.
       A group-decision-network, which is supervised by the self-distributed labeling, outputs a set of group probabilities $\left\{ p(G_0|\mathbf{x}), p(G_1|\mathbf{x}), ..., p(G_{K-1}|\mathbf{x}) \right\} $ from the instance-based representation.
       The final representation of 512 dimension is an aggregation of the instance-based representation and the weighted sum $\mathbf{v}_{\mathbf{x}}^{{G}}$ of the group-aware representations with the group probabilities. 
       W is a weight of the function $g$.
       }
    \label{fig:network}
\end{figure*}

\section {Proposed Method}

Our GroupFace learns the latent groups by using a self-distributed grouping method, constructs multiple group-aware representations and ensembles them into the standard instance-based representation to enrich the feature representation for face recognition.

\subsection {GroupFace}
We discuss that how the scheme of latent groups are effectively integrated into the embedding feature in GroupFace. 
\smallbreak

\noindent\textbf{Instance-based Representation.}
We will call a feature vector in conventional face recognition scheme \cite{ArcFace, CosFace, centerloss, regularface} as an {\it Instance-based Representation} in this paper (Figure~\ref{fig:network}).
The instance-based representation is commonly trained as an embedding feature by using softmax-based loss (\eg, CosFace \cite{CosFace} and ArcFace \cite{ArcFace}) and is used to predict an identity as:
\begin{equation}
    p(\mathbf{y}_i|\mathbf{x}) = \text{softmax}_k(g(\textbf{v}_\textbf{x})),
\end{equation}
where $\mathbf{y}_i$ is an identity label, $\textbf{v}_\textbf{x}$ is the instance-based representation of a given sample $\textbf{x}$, and $g$ is a function which projects an embedding feature of 512 dimension into $M$ dimensional space. $M$ is the number of person identities.
\smallbreak

\noindent\textbf{Group-aware Representation.}
GroupFace uses a novel {\it Group-aware Representation} as well as the instance-based representation to enrich the embedding features.
Each group-aware representation vector is extracted by deploying fully-connected layers for each corresponding group (Figure \ref{fig:network}).
The embedding feature ($\bar{\mathbf{v}}_{\mathbf{x}}$, \textit{Final Representation} in Figure \ref{fig:network}) of GroupFace is obtained by aggregating the instance-based representation $\textbf{v}_\textbf{x}$ and the weighted-summed group-aware representation $\textbf{v}^{G}_\textbf{x}$.
GroupFace predicts an identity by using the enriched final representation $\bar{\mathbf{v}}_{\mathbf{x}_i}$ as:
\begin{equation}
\begin{aligned}
    p(\mathbf{y}_i|\mathbf{x}) 
    &= \text{softmax}_k(g(\bar{\mathbf{v}}_{\mathbf{x}})),\\
    &= \text{softmax}_k(g(\textbf{v}_\textbf{x}+\textbf{v}^{{G}}_\textbf{x})),
    \label{eq:groupaware}
\end{aligned}
\end{equation}
where $\textbf{v}^{{G}}_\textbf{x}$ is an ensemble of multiple group-aware representations with group probabilities.
\smallbreak

\noindent\textbf{Structure.} 
GroupFace calculates and uses instance-based representation and group-aware representations, concurrently.
The instance-based representation is obtained by the same procedures that are used in conventional face recognition methods \cite{ArcFace, CosFace, centerloss, regularface}, and the $K$ group-aware representations are obtained similarly by deploying a fully-connected layer.
Then, the group probabilities are calculated from the instance-based representation vector by deploying a Group Decision Network (GDN) that consists of three fully-connected layers and a softmax layer.
Using the group probabilities, the multiple group-aware representations are sub-ensembled in a soft manner (S-GroupFace) or a hard manner (H-GroupFace).
\begin{enumerate}
    \item 
        S-GroupFace aggregates multiple group-aware representations with corresponding probabilities of groups as weights, and is defined as:
        \begin{equation}
            \textbf{v}^{{G}}_\textbf{x} = \sum _ {k \in K} p({G}_k|\mathbf{x}){\textbf{v}^{{G}_k}_\textbf{x}}.
        \end{equation}
    \item 
        H-GroupFace selects one of the group-aware representations for which the corresponding group probability has the highest value, and is defined as:
        \begin{equation}
            \textbf{v}^{{G}}_\textbf{x} = \operatorname*{arg\,max}_{p({G}_k|\mathbf{x})} {\textbf{v}^{{G}_k}_\textbf{x}}.
        \end{equation}
\end{enumerate}
S-GroupFace provides a significant improvement of recognition accuracy with a marginal requirement for additional resources, and H-GroupFace is more suitable for practical applications than S-GroupFace, at the cost of a few additional convolutions.
The final representation $\bar{\textbf{v}}_{\mathbf{x}}$ is enriched by aggregating both the instance-based representation and the sub-ensembled group-aware representation.
\smallbreak

\noindent\textbf{Group-aware Similarity.}
We introduce a group-aware similarity that is a new similarity considering both the standard embedding feature and the intermediate feature of GDN in the inference stage.
The group-aware similarity is penalized by a distance between intermediate features of two given instances because the intermediate feature is not trained on the cosine space and just describes the group identity of a given sample, not the explicit identity of a given sample.
The group-aware similarity $S^*$ between the $i^{th}$ image $I_i$ and the $j^{th}$ image $I_j$ is defined as:
\newcommand{\norm}[1]{\left\lVert#1\right\rVert}
\begin{equation}
    \label{eq:group_similarity}
    S^* (\mathbf{x}_i, \mathbf{x}_j) = S (\bar{\mathbf{v}}_{\mathbf{x}_i}, \bar{\mathbf{v}}_{\mathbf{x}_j} ) - \beta D(\hat{\mathbf{v}}_{\mathbf{x}_i}, \hat{\mathbf{v}}_{\mathbf{x}_j} )^{\gamma},
\end{equation}
where $S$ is a cosine similarity metric, $D$ is a distance metric, $\hat{\mathbf{v}}_{\mathbf{x}}$ denotes the intermediate feature of GDN and, $\beta$ and $\gamma$ are a constant parameter. 
The parameters are determined empirically to be $\beta = 0.1$ and $\gamma = 1/3$.

\subsection {Self-distributed Grouping}
In this work, we define a group as a set of samples that share any common visual-or-non-visual features that are used for face recognition.
Such a group is determined by a deployed GDN.
Our GDN is gradually trained in a self-grouping manner that provides a group label by considering the distribution of latent groups without any explicit ground-truth information.
\smallbreak


\noindent\textbf{Na\"ive Labeling. }
A na\"ive way to determine a group label is to take an index that has the maximum activation of softmax outputs.
We build a GDN $f$ to determine a belonging group $G^*$ for a given sample $\mathbf{x}$ by deploying MLP and attaching a softmax function:
\begin{equation}
    p(G_k|\mathbf{x}) = \text{softmax}_k(f(\mathbf{x})),
    \label{eq:naiveway1}
\end{equation}
\begin{equation}
    G^* (\mathbf{x}) = 
    \operatorname*{arg\,max}_k p(G_k|\mathbf{x}),
    \label{eq:naiveway2}
\end{equation}
where $G_k$ is the $k^{th}$ group.
The lack of the consideration for the group distribution can cause the na\"ive solution to assign most of samples to one or few groups.
\smallbreak

\begin{figure}[t]
    \begin{center}
        \includegraphics[width=8.4cm]{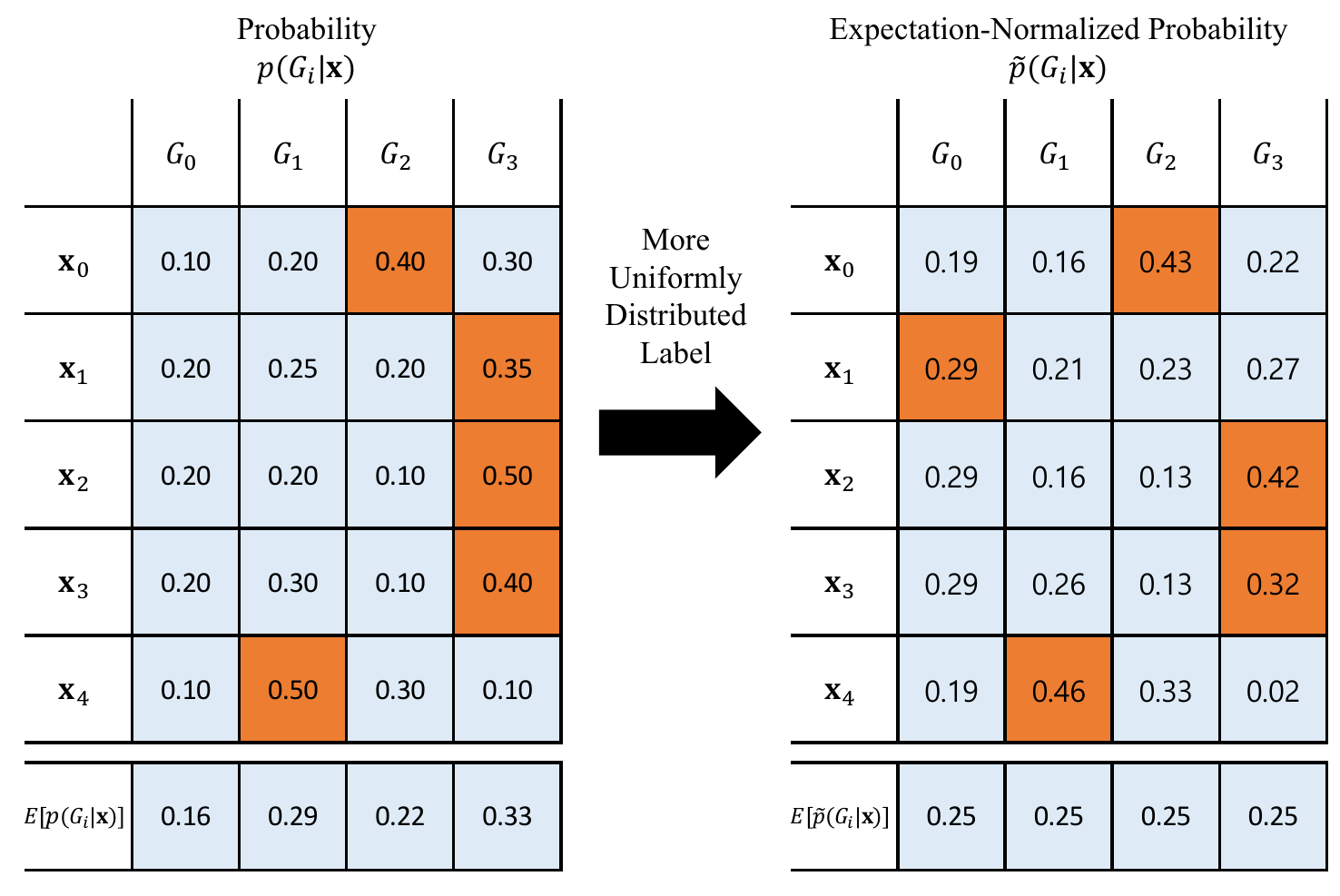}
    \end{center}
    \caption{Conceptual scheme of self-distributed labeling. 
    When group probabilities of 5 samples on 4 groups is obtained by GDN, the expectation-normalized probability is computed by subtracting the corresponding expectation of the group probabilities from the group probabilities. 
    The group identifier of the maximum probability is selected as the label of a given sample and the consideration of the expectation makes the distribution of labels more uniform.}
    \label{fig:vcluster}
\end{figure}

\noindent\textbf{Self-distributed Labeling.}
We introduce an efficient labeling method that utilizes a modified probability regulated by a prior probability to generate uniformly-distributed group labels in a deep manner.
We define an expectation-normalized probability $\tilde{p}$ to balance the number of samples among $K$ groups:
\begin{equation}
    \tilde{p}(G_k|\mathbf{x}) = \frac{1}{K} \left\{ p(G_k|\mathbf{x}) - E_{\mathbf{x} \sim \text{data}} \left [ {p}({G}_k|\mathbf{x}) \right ] \right\} + \frac{1}{K},
\end{equation} 
where the first $1/K$ bounds the normalized probability between 0 and 1.
Then, the expectation of the expectation-normalized probability is computed as:
\begin{equation}
\begin{aligned}
    &E_{\mathbf{x} \sim \text{data}} \left [ \tilde{p}({G}_k|\mathbf{x}) \right ], \\
    &= \frac{1}{K} \left \{ E_{\mathbf{x} \sim \text{data}} \left [ {p}({G}_k|\mathbf{x}) \right ] - E_{\mathbf{x} \sim \text{data}} \left [ {p}({G}_k|\mathbf{x}) \right ] \right \} + \frac{1}{K},   \\
    &= \frac{1}{K}.
\end{aligned}
\end{equation}
The optimal self-distributed label is obtained as:
\begin{equation}
    {G}^* (\mathbf{x}) = 
    \operatorname*{arg\,max}_k \tilde{p}({G}_k|\mathbf{x}).
\end{equation}
The trained GDN estimates a set of group probabilities that represent the degree to which the sample belongs to the latent groups.
As the number of samples approaches infinity, the proposed method stably outputs the uniform-distributed labels (Figure \ref{fig:vcluster}).
\smallbreak

\subsection{Learning}
The network of GroupFace is trained by 
both the standard classification loss, which is a softmax-based loss to distinguish identities,
and the self-grouping loss, which is a softmax loss to train latent groups, simultaneously.
\smallbreak

\noindent\textbf{Loss Function. } 
A softmax-based loss $L_{1}$ (ArcFace \cite{ArcFace} is mainly used in this work) is used to train a feature representation for identities and is defined as:
\begin{equation}
    L_{1} = - \frac{1}{N} \sum _{i=1} ^ {N} \log { \frac{ e^{s(\cos(\theta_{\mathbf{y}_i} + m ))} }{ e^{s(\cos(\theta_{\mathbf{y}_i} + m ))} + \sum _{j=1, j \neq \mathbf{y}_i} ^{n} e^{s(\cos(\theta_{j} ))} } },
\end{equation}
where $N$ is a number of samples in a mini-batch, $\theta$ is the angle between a feature and the corresponding weight, $s$ is a scale factor, $m$ is a marginal factor.
To construct the optimal group-space, a self-grouping loss, which reduces the difference between the prediction and the self-generated label, is defined as:
\begin{equation}
    L_{2} = - \frac{1}{N}  \sum _{i=1} ^ {N} \text{CrossEntropy} (\text{softmax} (f(\mathbf{x}_i)), G^* (\mathbf{x}_i)).
    \end{equation}
\smallbreak

\noindent\textbf{Training.}
The whole network is trained by using the aggregation of two losses:
\begin{equation}
    L=L_{1} + \lambda L_{2},
\end{equation}
where the parameter $\lambda$ balances the weights of different losses and is empirically set to $0.1$.
Thus, GDN can learn the group, which is an attribute beneficial to face recognition.
\smallbreak

\section{Experiments}

We describe implementation details and extensively perform experiments and ablation studies to show the effectiveness of GroupFace.

\subsection{Implementation Details}

\noindent\textbf{Datasets.}
For the train, we use MSCeleb-1M \cite {msceleb1m} which has contain about 10M images for 100K identities.
Due to the noisy labels of MSCeleb-1M original dataset, we use the refined version \cite {ArcFace} which contains 3.8M images for 85k identities.
For the test, we conduct our experiments with nine commonly used datasets as follows: 
\begin{itemize}
\item LFW \cite{LFW} which contains 13,233 images from 5,749 identities and provides 6000 pairs from them. CALFW \cite{CALFW} and CPLFW \cite{CPLFWTech} are the reorganized datasets from LFW to include higher pose and age variations. 
\item YTF \cite{YTF} which consists of 3,425 videos of 1,595 identities.
\item MegaFace \cite{kemelmacher2016megaface} which composed of more than 1 million images from 690K identities for challenge 1(MF1).
\item CFP-FP \cite{CFP_FP} which contains 500 subjects, each with 10 frontal and 4 profile images.
\item AgeDB-30 \cite{age_db} which contains 12,240 images of 440 identities.
\item IJB-B \cite{IJB_B} which contains 67,000 face images, 7,000 face videos and 10,000 non-face images.
\item IJB-C \cite{IJB_C} which contains 138,000 face images, 11,000 face videos and 10,000 non-face images.
\end{itemize}

\noindent\textbf{Metrics.}
We compare the verification-accuracy for identity-pairs on LFW \cite{LFW}, YTF \cite{YTF}, CALFW \cite{CALFW}, CPLFW \cite{CPLFWTech}, CFP-FP \cite{CFP_FP}, AgeDB-30 \cite{age_db} and MegaFace \cite{kemelmacher2016megaface} verification task.
MegaFace \cite{kemelmacher2016megaface} identification task is evaluated by rank-1 identification accuracy with 1 million distractors.
We compare a True Accept Rate at a certain False Accept Rate (TAR@FAR) from 1e-4 to 1e-6 on IJB-B \cite{IJB_B} and IJB-C \cite{IJB_C}.
\smallbreak

\noindent\textbf{Experimental Setting.}
We construct a normalized face image \cite{ArcFace, SphereFace, CosFace} ($112 \times 112$) by warping a face-region using five facial points from two eyes, nose and two corners of mouth.
We employ the ResNet-100 \cite{resnet} as the backbone network similar to the recent works \cite{ArcFace, AFRN}.
We vectorize the activation and reduced \# activations to 4096 (shared feature in Figure \ref{fig:network}) by a block of BN-FC. 
Our GroupFace is attached after \textit{res5c} in ResNet-100, where its activation dimension is 512$\times$7$\times$7.
The MLP in GDN consists of two blocks of BN-FC and a FC for group classification.
We follow \cite{ArcFace, CosFace} to set the hyper-parameters of the loss function.
\smallbreak


\begingroup
\newcolumntype{C}[1]{>{\centering\let\newline\\\arraybackslash\hspace{0pt}}m{#1}}
\begin{table}[t!]
	\begin{center}
		\centering
		\subfloat[Number of Groups]
		{
			\centering
			\begin{tabular}[t]{p{2.0cm}|C{2.2cm}C{2.2cm}}
				\hline 
				&    \multicolumn{2}{c}{ {TAR} } \\
				&    {FAR=1e-6} & {FAR=1e-5} \\
				\hline 
				\hline 
				{Baseline \cite{ArcFace}}	& 0.3828 & 0.8933\\
    			{- 4 Groups}	    & 0.4395 & 0.8962 \\
				{- 16 Groups}	    & 0.4435 & 0.8993\\
			    {- 32 Groups}	    & 0.4678 & 0.9115 \\
				\hline 
			\end{tabular}
			\label{tab:table1a}
		} \\
		\subfloat[Learning for GDN]
		{
			\centering
			\begin{tabular}[t]{p{4.0cm}|C{2.8cm}}
				\hline 
				&    {TAR@FAR=1e-6}  \\
				\hline 
				\hline 
				{Baseline \cite{ArcFace}}		 	  & {0.3828}  \\
				{- without Loss}	  & {0.4468}  \\
				{- Naive Labeling}	  & {0.4535}  \\
				{- Self-distributed Labeling }	  & {0.4678}\\
				\hline 
			\end{tabular}
			\label{tab:table1b}
		} \\
		\subfloat[Hard \textit{vs.} Soft Ensemble]
		{
			\centering
			\begin{tabular}[t]{p{2.2cm}|C{2.8cm}C{1.4cm}}
				\hline 
				&    {TAR@FAR=1e-6} & {GFLOPS}   \\
				\hline 
				\hline 
				{Baseline \cite{ArcFace}}		  & {0.3828} & {24.2G} \\
				{- H-GroupFace}	  & {0.4439} & {24.4G} \\
				{- S-GroupFace}	  & {0.4678} & {24.5G} \\
				\hline 
			\end{tabular}
			\label{tab:table1c}
		} \\
		\subfloat[Aggregation vs. Concatenataion]
		{
			\centering
			\begin{tabular}[t]{p{2.0cm}|C{2.2cm}C{2.2cm}}
				\hline 
				&    \multicolumn{2}{c}{ {TAR} } \\
				&    {FAR=1e-6} & {FAR=1e-5} \\
				\hline 
				\hline 
				{Baseline \cite{ArcFace}}		  & 0.3828 & 0.8933 \\
				{- Concat} & 0.4745 & 0.8999 \\
				{- Aggregation}	  & 0.4678 & 0.9115 \\
				\hline 
			\end{tabular}
			\label{tab:table1d}
		} \\
		\subfloat[Group-aware Similarity]
		{
			\centering
			\begin{tabular}[t]{p{4.0cm}|C{2.8cm}}
				\hline 
				&    {TAR@FAR=1e-6}  \\
				\hline 
				\hline 
				{Baseline \cite{ArcFace}}		  & 0.3828 \\
				{- GroupFace}	  & 0.4678 \\
				{- GroupFace$^{\dag}$}    & 0.5212 \\
				\hline 
			\end{tabular}
			\label{tab:table1e}
		} \\
		\subfloat[Lightweight Model]
		{
			\centering
			\begin{tabular}[t]{p{2.0cm}|C{1.3cm}C{1.3cm}C{1.3cm}}
				\hline 
				&    \multicolumn{3}{c}{\scalebox{0.9}{TAR}}  \\
				&    \scalebox{0.8}{FAR=1e-6} & \scalebox{0.8}{FAR=1e-5} & \scalebox{0.8}{FAR=1e-4}  \\
				\hline 
				\hline 
				\scalebox{0.9}{ResNet-100 \cite{ArcFace} }      & 0.3828 & 0.8933 & 0.9425 \\
				\scalebox{0.9}{ResNet-34  \cite{ArcFace} }	    & 0.3962 & 0.8669 & 0.9308 \\
				\scalebox{0.9}{+ GroupFace}	                    & 0.4361 & 0.8820 & 0.9316 \\
				\scalebox{0.9}{+ GroupFace$^{\dag}$}            & 0.4823 & 0.8842 & 0.9354 \\
				\hline 
			\end{tabular}
			\label{tab:table1f}
		}
	\end{center}
	\caption{
		Ablation studies for the proposed GroupFace on IJB-B dataset. The baseline is a recognition-model trained by ArcFace \cite{ArcFace} and $^{\dag}$ denotes an evaluation procedure using the group-aware similarity (Eq. \ref{eq:group_similarity}).
	}
\end{table}
\endgroup

\noindent\textbf {Learning.}
We train the model with $8$ synchronized GPUs and a mini-batch involving $128$ images per GPU.
To stable the group-probability, the network of GroupFace is trained from the pre-trained network that is trained by only the softmax-based loss \cite{ArcFace, CosFace}.
We used a learning rate of $0.005$ for the first 50k, $0.0005$ for the 20k, and $0.00005$ for 10k with a weight decay of $0.0005$ and a momentum of $0.9$ with stochastic gradient descent (SGD).
We compute the expectation of group probabilities by computing the group probabilities of $128 \times 8$ samples on all GPUs and averaging the expectations over the recent $B$-batches to accurately estimate the expectation of the group probabilities on the population; $B$ between $32$ and $128$ empirically shows a similar performance.


\subsection{Ablation Studies}
To show the effectiveness of the proposed method, we perform the ablation studies on the it's behaviors.
For all experiments, we also use the same network structure with the hyper-parameters mentioned earlier.
To clearly show the effect of each ablation study, TAR@FAR of the models are compared on IJB-B dataset \cite{IJB_B}; all models in the ablation studies shows around $99.85\%$ on LFW.
\smallbreak

\noindent\textbf{Number of Groups.}
We compare the recognition performance according to the number of groups (Table \ref{tab:table1a}).
As the number of groups grows, the performance increases steadily.
In particular, a few initial groups can benefit greatly, and by deploying more groups, significant improvement of performance can be obtained.
\smallbreak

\noindent\textbf{Learning for GDN.}
We compare the learning method for GDN (Table \ref{tab:table1b}): (1) without loss (adopt the group-aware network structure only), (2) naive labeling, and (3) self-distributed labeling.
Just by applying our novel network structure, the recognition performance is greatly improved.
In particular, the performance is further increased by adjusting the proposed self-distributed labeling method.
\smallbreak

\noindent\textbf{Hard \textit{vs.} Soft.}
S-GroupFace shows a high improvement in the performance because it uses all group-aware representations comprehensively with a reasonable additional resource (Table \ref{tab:table1c}).
Since H-GroupFace uses only one strongest group-aware representation even if many groups are deployed, the burden of increasing the number of groups is fixed to a slight amount of additional resource.
Thus, H-GroupFace can be applied immediately for high performance gains in practical applications.
\smallbreak

\noindent\textbf{Aggregation \textit{vs.} Concatenation.}
We compare how to combine the instance-based representation and the group-aware representations into an one embedding feature (Table \ref{tab:table1d}): (1) aggregation and (2) concatenation.
Concatenation-based GroupFace shows a better TAR@FAR=1e-6 by 0.67 percentage points than Aggregation-based GroupFace, 
however, Aggregation-based GroupFace shows a much better TAR@FAR=1e-5 by 1.16 percentage points.
We chose the Aggregation-based GroupFace that is generally better performing with fewer feature dimensions.
\smallbreak

\noindent\textbf{Group-aware Similarity.}
The recognition-performance is once again improved significantly by evaluating the group-aware similarity (Table \ref{tab:table1e}).
Even though the group-aware similarity increases the feature dimension for calculating a similarity, it is easy to extract the required feature because the feature is the intermediate output of the recognition network.
Especially, this experiment shows that the group-based information is distinct from the conventional identity-based information enough to improve performance in practical usages.
We show more detailed experiments in Table \ref{tab:table5}.
\smallbreak

\noindent\textbf{Lightweight Model.}
GroupFace is also effective for a lightweight model such as ResNet-34 \cite{resnet} that requires only 8.9 GFLOPS less than ResNet-100 \cite{resnet}, which requires 24.2 GFLOPS.
ResNet-34 based GroupFace shows a similar performance of ResNet-100 based ArcFace \cite{ArcFace} and greatly outperforms ResNet-100 in a most difficult criterion (FAR=1e-6).
In addition, the group-aware similarity significantly exceed the basic performance of ResNet-34 model (Table \ref{tab:table1f}).
\smallbreak

\subsection{Evaluations}

 \begin{table}
    \begin{center}
    \begin{tabular}{cccc}
    \hline
    Method & \#Image & LFW & YTF \\
    \hline
    \hline
    DeepID \cite{contrastive_loss}      & 0.2M & 99.47 & 93.2 \\
    DeepFace \cite{deepface}            & 4.4M & 97.35 & 91.4 \\
    VGGFace \cite{vggface}              & 2.6M & 98.95 & 97.3 \\
    FaceNet \cite{FaceNet}              & 200M & 99.64 & 95.1 \\
    CenterLoss \cite{centerloss}        & 0.7M & 99.28 & 94.9 \\
    RangeLoss \cite{range_loss}         & 5M   & 99.52 & 93.7 \\
    MarginalLoss \cite{marginal_loss}   & 3.8M & 99.48 & 95.9 \\
    SphereFace \cite{SphereFace}        & 0.5M & 99.42 & 95.0 \\
    RegularFace \cite{regularface}      & 3.1M & 99.61 & 96.7 \\
    CosFace \cite{CosFace}              & 5M   & 99.81 & 97.6 \\
    UniformFace \cite{uniformface}      & 6.1M & 99.80 & 97.7 \\
    AFRN \cite{AFRN}                    & 3.1M & 99.85 & 97.1 \\
    ArcFace \cite{ArcFace}              & 5.8M & 99.83 & 97.7 \\
    GroupFace                           & 5.8M & 99.85 & 97.8 \\
    \hline
    \end{tabular}
    \end{center}
     \caption{Verification accuracy ($\%$) on LFW and YTF.}
			\label{tab:table2}
 \end{table}
 
\begin{table}
    \begin{center}
    \begin{tabular}{ccccc}
    \hline
    \scalebox{0.8}{Method} & \scalebox{0.8}{CALFW} & \scalebox{0.8}{CPLFW} & \scalebox{0.8}{CFP-FP} & \scalebox{0.8}{AgeDB-30} \\
    \hline
    \hline
    CenterLoss \cite{centerloss}    & 85.48 & 77.48 & - & -\\
    SphereFace \cite{SphereFace}    & 90.30 & 81.40 & - & -\\
    VGGFace2 \cite{vggface2}        & 90.57 & 84.00 & - & -\\
    CosFace \cite{CosFace}          & 95.76 & 92.28 & 98.12 & 98.11\\
    ArcFace \cite{ArcFace}          & 95.45 & 92.08 & 98.27 & 98.28 \\
    GroupFace                       & 96.20 & 93.17 & 98.63 & 98.28\\
    \hline
    \end{tabular}
    \end{center}
     \caption{Verification accuracy ($\%$) on CALFW, CPLFW, CFP-FP and AgeDB-30.}
			\label{tab:table3}
 \end{table}

\begin{table}
    \begin{center}
    \begin{tabular}{cccc}
    \hline
    \scalebox{0.8}{Method} & \scalebox{0.8}{Protocol} & \scalebox{0.8}{Ident} & \scalebox{0.8}{Verif} \\
    \hline
    \hline
    RegularFace \cite{regularface}      & Large & 75.61 & 91.13 \\
    UniformFace \cite{uniformface}      & Large & 79.98 & 95.36 \\
    CosFace \cite{CosFace}              & Large & 80.56 & 96.56 \\
    ArcFace \cite{ArcFace}              & Large & 81.03 & 96.98 \\
    GroupFace                           & Large & 81.31 & 97.35 \\
    \hline
    SphereFace \cite{SphereFace}        & Large / R & 97.91 & 97.91 \\
    AdaptiveFace \cite{adaptiveface}    & Large / R & 95.02 & 95.61 \\
    CosFace \cite{CosFace}              & Large / R & 97.91 & 97.91 \\
    ArcFace \cite{ArcFace}              & Large / R & 98.35 & 98.49 \\
    GroupFace                           & Large / R & 98.74 & 98.79 \\
    \hline
    \end{tabular}
    \end{center}
     \caption{Identification and verification evaluation on MegaFace.
     Ident means rank-1 identification accuracy ($\%$) and Verif means TAR@FAR=1e-6 ($\%$).
     R denotes the evaluation procedure on the refined version \cite{ArcFace} of MegaFace dataset.
     }
			\label{tab:table4}
 \end{table}

\begin{table*}[ht!]
   \begin{center}
   \begin{tabular}{p{3.5cm}|p{1.8cm}p{1.8cm}p{1.8cm}|p{1.8cm}p{1.8cm}p{1.8cm}}
   \hline
   \multirow{2}{*}{Method} &  \multicolumn{3}{c}{TAR on IJB-B} &  \multicolumn{3}{c}{TAR on IJB-C}  \\
   \cline{2-4}\cline{5-7}
    & {FAR=1e-6} & {FAR=1e-5} & {FAR=1e-4} & {FAR=1e-6} & {FAR=1e-5} & {FAR=1e-4}  \\
    \hline\hline
    VGGFace2 \cite{vggface2}                & - & 0.671 & 0.800 &  - & 0.747 & 0.840 \\
    CenterFace \cite{centerloss}            & - & - & - & - & 0.781 & 0.853 \\
    ComparatorNet \cite{ComparatorNet}           & - & -     & 0.849 &  - & - & 0.885 \\
    PRN \cite{PRN}                                 & - & 0.721 & 0.845 &  - & - & - \\
    AFRN \cite{AFRN}                    & - & 0.771 & 0.885 &  - & 0.884 & 0.931 \\
    CosFace \cite{CosFace}  & 0.3649 & 0.8811 & 0.9480 &  0.8801 & 0.9370 & 0.9615 \\
    ArcFace \cite{ArcFace}  & 0.3828 & 0.8933 & 0.9425 &  0.8625 & 0.9315 & 0.9565 \\
    GroupFace$^{-}$           & 0.4166 & 0.8983 & 0.9453 &  0.8858 & 0.9399 & 0.9606\\
    GroupFace               & 0.4678 & 0.9115 & 0.9445 &  \textbf{0.9053} & 0.9437 & 0.9602\\
    GroupFace$^{\dag}$      & \textbf{0.5212} & \textbf{0.9124} & \textbf{0.9493} &  0.8928 & \textbf{0.9453} & \textbf{0.9626}\\
   \hline
   \end{tabular}
   \end{center}
    \caption{Verification evaluation according to different FARs on IJB-B and IJB-C.
    GroupFace is trained by ArcFace \cite{ArcFace}.
    $^{-}$ denotes that a model is trained by CosFace \cite{CosFace} and $^{\dag}$ denotes that a model is evaluated by using the group-aware similarity.}
    \label{tab:table5}
\end{table*}

\noindent\textbf{LFW, YTF, CALFW, CPLFW, CFP-FP}\textbf{ and AgeDB-30.}
We compare the verification-accuracy on LFW \cite{LFW} and YTF \cite{YTF} with the unrestricted with labelled outside data protocol (Table \ref{tab:table2}).
On YTF, we evaluate all the images without the exclusion of noisy images from image sequences.
Even though both datasets are highly-saturated, Our GroupFace surpasses the other recent methods.
We also report the verification accuracy on the variant of LFW (CALFW \cite{CALFW}, CPLFW \cite{CPLFWTech}), CFP-FP \cite{CFP_FP} and AgeDB-30 \cite{age_db} (Table \ref{tab:table3}).
Our GroupFace shows the better accuracy on all of the above datasets.
\smallbreak

\noindent\textbf{MegaFace.}
We evaluate our GroupFace under the large-training-set protocol, in which models are trained by using the training set containing more than 0.5M images, on MegaFace \cite{kemelmacher2016megaface} (Table \ref{tab:table4}).
GroupFace is the top-ranked face recognition model among the recent published state-of-the-art methods.
On the refined MegaFace \cite{ArcFace}, our GroupFace also outperforms the other models.
\smallbreak

\noindent\textbf{IJB-B and IJB-C.}
We compare the proposed method with other methods on IJB-B \cite{IJB_B} and IJB-C \cite{IJB_C} datasets (Table \ref{tab:table5}).
Recent angular-margin-softmax based methods \cite{ArcFace, CosFace} show great performance in the datasets.
We reports the improvement of GroupFace in the verification accuracy based on both CosFace \cite{CosFace} and ArcFace \cite{ArcFace} without any test-time augmentations such as horizontal flipping.
Our GroupFace shows significant improvements on all FAR criteria by 8.5 percentage points on FAR=1e-6, 1.8 percentage points on FAR=1e-5 and 0.2 percentage points on FAR=1e-4 than the ArcFace \cite{ArcFace} on IJB-B and by 4.3 percentage points on FAR=1e-6, 1.2 percentage points on FAR=1e-5 and 0.4 percentage points on FAR=1e-4 than the ArcFace \cite{ArcFace} on IJB-C. 
The recognition-performance is once again improved significantly by applying the group-aware similarity (Eq. \ref{eq:group_similarity}), especially on the most difficult criterion (TAR@FAR=1e-6) on IJB-B by 5.3 percentage points.

\subsection{Visualization}

To show the effectiveness of the proposed method, we visualize the feature representation, the average activation of groups and the visual interpretation of groups.
\\

\begin{figure}[t]
    \begin{center}
		\subfloat[Baseline]{		
			\label{fig:va1}
            \fbox{\includegraphics[width=2.4cm]{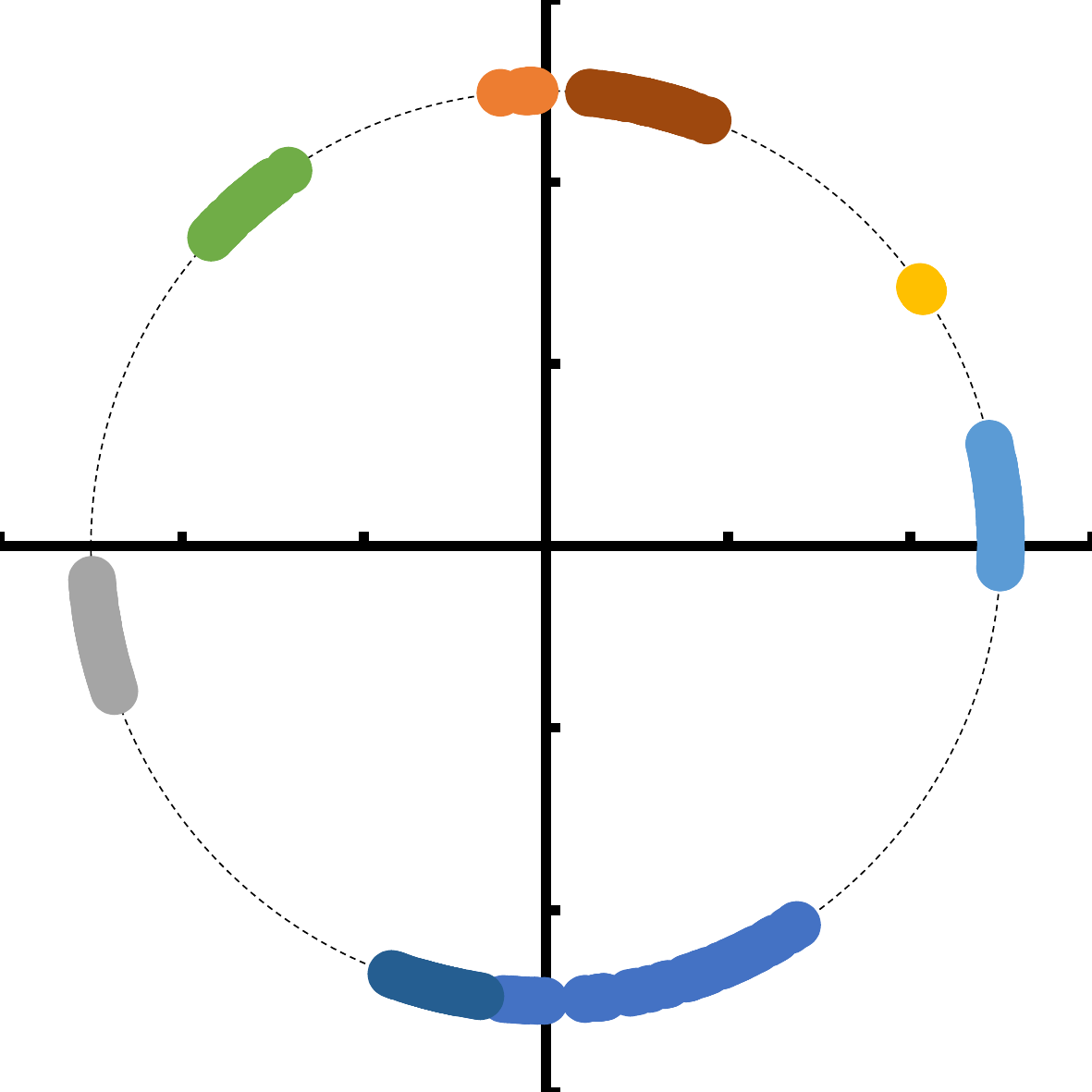}}
		}	
		\subfloat[$\mathbf{v}_\mathbf{x}$]{
			\label{fig:rb2}
            \fbox{\includegraphics[width=2.4cm]{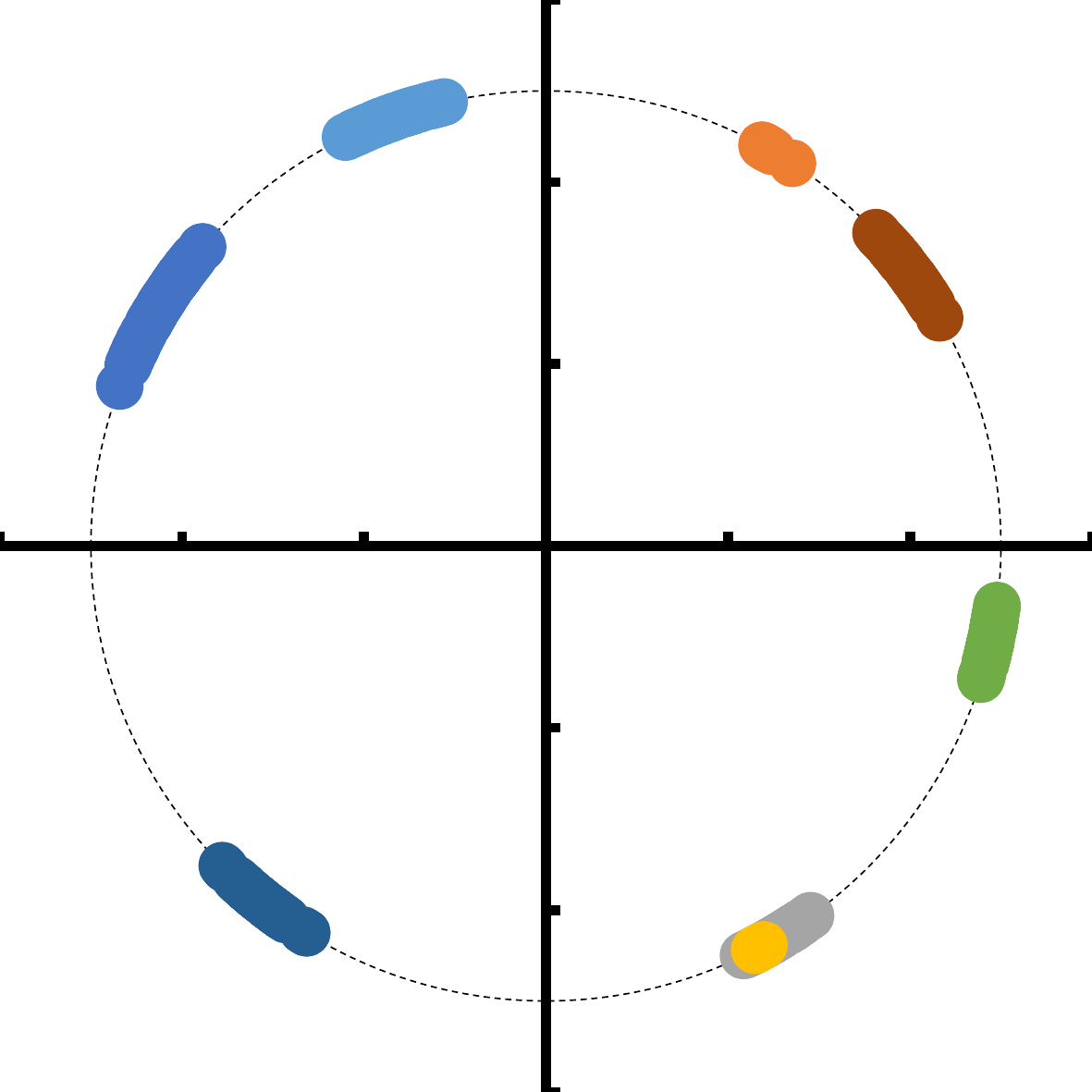}}
		}	
		\subfloat[$\bar{\mathbf{v}}_\mathbf{x}$]{
			\label{fig:va2}
            \fbox{\includegraphics[width=2.4cm]{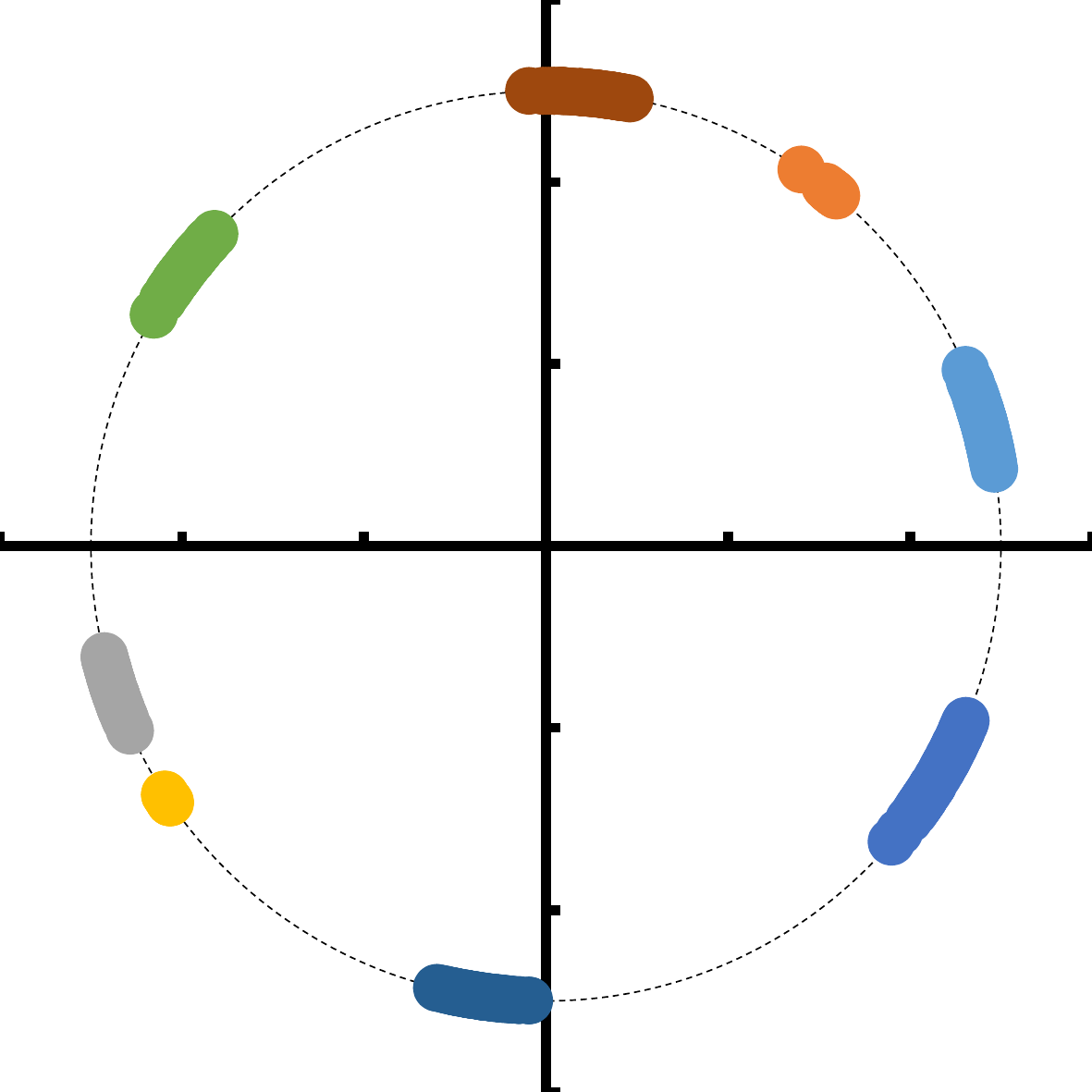}}
		}	
    \end{center}
    \caption{
        A quantitative comparison of the baseline network (ArcFace \cite{ArcFace}), $\mathbf{v}_\mathbf{x}$ that denotes an instance-based representation of GroupFace and $\bar{\mathbf{v}}_\mathbf{x}$ that denotes a final representation of GroupFace for the first eight identities in the refined MSCeleb-1M \cite{msceleb1m} dataset on a 2D space.
        The eight colored-circles represents the eight identities.}
    \label{fig:vis_angle}
\end{figure}

\begin{figure}[t]
    \begin{center}
    \includegraphics[width=8cm]{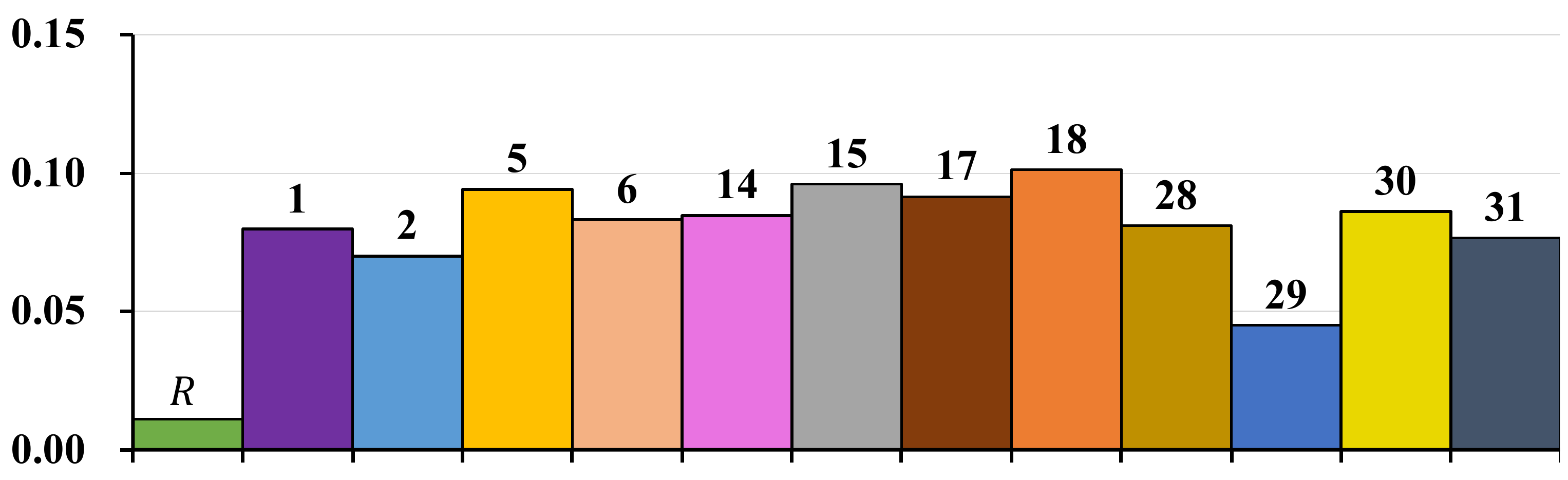}
    \end{center}
   \caption{Average probabilities of 32 groups on the refined MSCeleb-1M \cite{msceleb1m} dataset.
   12 groups can mainly activate and the rest groups, $R$, softly affect. }
\label{fig:activationprob}
\end{figure}

\noindent\textbf{2D Projection of Representation.}
Figure~\ref{fig:vis_angle} shows a quantitative comparison among (a) the final representation of the baseline network (ArcFace \cite{ArcFace}), (b) the instance-based representation of GroupFace and the final representation of GroupFace on a 2D space.
We select the first eight identities in the refined MSCeleb-1M dataset \cite{msceleb1m} and map the extracted features onto the angular space by using t-SNE \cite{t_SNE}.
The quantitative comparison shows that the proposed model generates more distinctive feature representations rather than the baseline model and also the proposed model enhances the instance-based representation.
\smallbreak


\noindent\textbf{Activation Distribution of Groups.}
The proposed Self-Grouping tries to make the samples evenly spread throughout the all groups, and at the same time, the softmax-based loss also simultaneously propagates gradients into GDN so that the identification works best.
Thus, the probability distribution is not exactly uniform (Figure \ref{fig:activationprob}).
Some probabilities of the groups are low and the others are high (\eg, 1, 2, 5, 6, 14, 15, 17, 18, 28, 29, 30, 31$^{th}$ groups).
The overall distribution is not uniform as we expected, but we see that there is no dominant one among the high activated group.
\smallbreak

\begin{figure}[t!]
    \begin{center}
		\includegraphics[width=8.4cm]{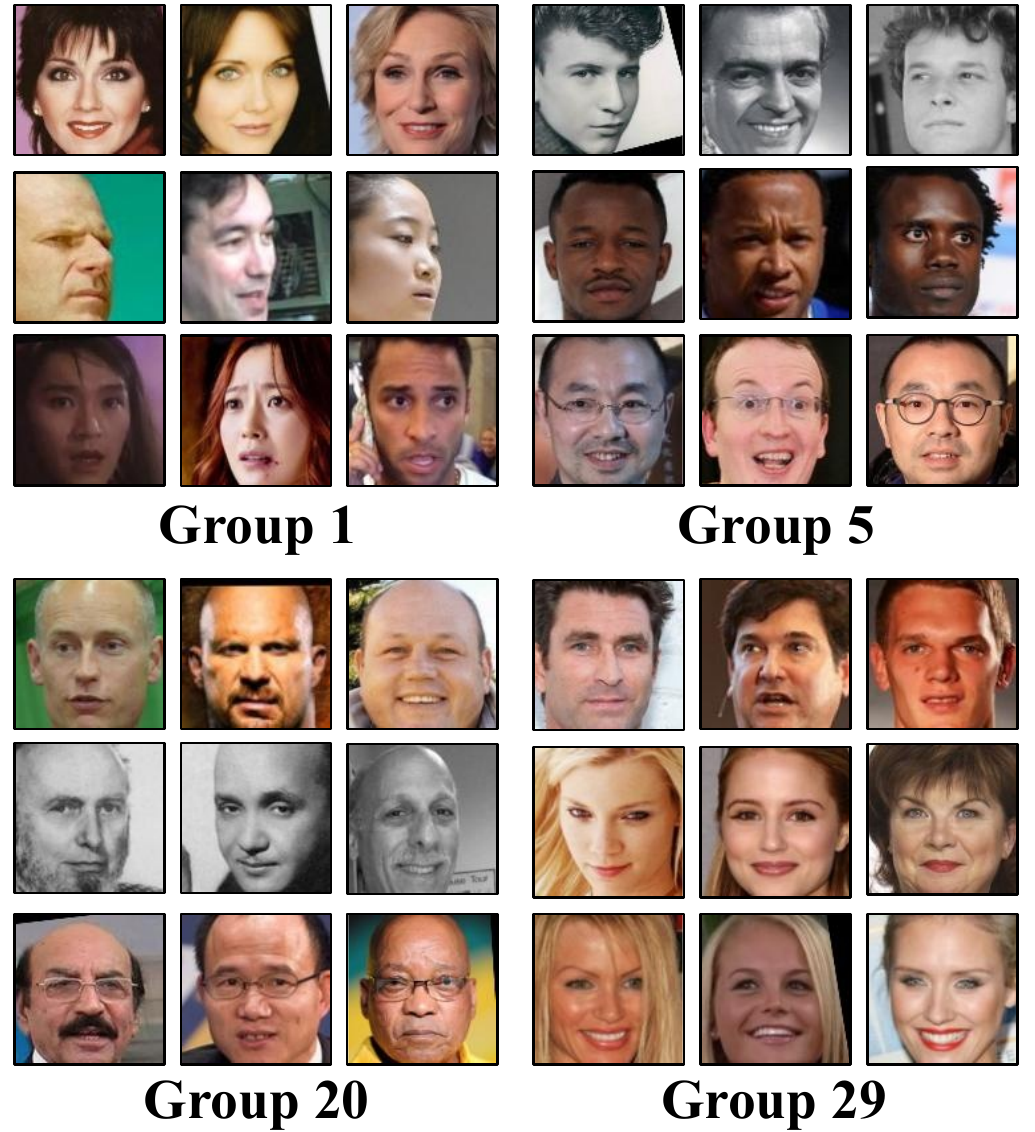}
    \end{center}
    \caption{
    Example images belonging to each groups. As enormous identities (80k$\sim$) of large scale dataset cannot be mapped to a few groups (32), each group contains identities of multiple characteristics.
    Some groups have one common visual description (Group 5: Some Men, Group 20: Bald Men) with some variations while others have multi-mode visual descriptions. 
    }
    \label{fig:attribute_visual}
\end{figure}

\noindent\textbf{Interpretation of Groups.}
The trained latent groups are not always visually distinguishable because they are categorized by a non-linear function of GDN using a latent feature, not a facial attribute (\eg, hair, glasses, and mustache).
However, there are two cases of groups (Group 5 and 20 in Figure~\ref{fig:attribute_visual}) that we can clearly see their visual properties; 95 of randomly-selected 100 images are men in Group 5 and 94 of randomly-selected 100 images are bald men in Group 20.
Others are not described as an one visual property, however, they seems to be described as multiple visual properties such as smile women, right-profile people and scared people in Group 1.

%

\section{Conclusion}
We introduce a new face-recognition-specialized architecture that consists of a group-aware network structure and a self-distributed grouping method to effectively manipulate multiple latent group-aware representations.
By extensively conducting the ablation studies and experiments, we prove the effectiveness of our GroupFace.
The visualization also shows that GroupFace fundamentally enhances the feature representations rather than the existing methods and the latent groups have some meaningful visual descriptions.
Our GroupFace provides a significant improvement in the recognition-performance ans is practically applicable to existing recognition systems. 
The rationale behind the effectiveness of GroupFace is summarized in two main ways:
(1) It is well known that additional supervisions from different objectives can bring an improvement of the given task by sharing a network for feature extraction, \eg, a segmentation head can improve accuracy in object detection~\cite{bell2016inside,he2017mask}. 
Likewise, learning the groups can be a helpful cue to train a more generalized feature extractor for face recognition.
(2) GroupFace proposes a novel structure that fuses instance-based representation and group-based representation, which is empirically proved its effectiveness.

\section*{Acknowledgement}
We thank AI team of Kakao Enterprise, especially Wonjae Kim and Yoonho Lee for their helpful feedback.

{\small
\bibliographystyle{ieee_fullname}
\bibliography{egpaper_for_review}

\begin{thebibliography}{10}\itemsep=-1pt

\bibitem{10.1007/978-3-540-24670-1_36}
Timo Ahonen, Abdenour Hadid, and Matti Pietik{\"a}inen.
\newblock Face recognition with local binary patterns.
\newblock In Tom{\'a}s Pajdla and Ji{\v{r}}{\'i} Matas, editors, {\em European
  Conference on Computer Vision Workshops}, 2004.

\bibitem{bell2016inside}
Sean Bell, C Lawrence~Zitnick, Kavita Bala, and Ross Girshick.
\newblock Inside-outside net: Detecting objects in context with skip pooling
  and recurrent neural networks.
\newblock In {\em IEEE Conference on Computer Vision and Pattern Recognition},
  2016.

\bibitem{vggface2}
Q. Cao, L. Shen, W. Xie, O.~M. Parkhi, and A. Zisserman.
\newblock Vggface2: A dataset for recognising faces across pose and age.
\newblock In {\em International Conference on Automatic Face and Gesture
  Recognition}, 2018.

\bibitem{caron2018deep}
Mathilde Caron, Piotr Bojanowski, Armand Joulin, and Matthijs Douze.
\newblock Deep clustering for unsupervised learning of visual features.
\newblock In {\em European Conference on Computer Vision}, 2018.

\bibitem{6619233}
D. {Chen}, X. {Cao}, F. {Wen}, and J. {Sun}.
\newblock Blessing of dimensionality: High-dimensional feature and its
  efficient compression for face verification.
\newblock In {\em IEEE Conference on Computer Vision and Pattern Recognition},
  2013.

\bibitem{joint_bayesian}
D. {Chen}, X. {Cao}, D. {Wipf}, F. {Wen}, and J. {Sun}.
\newblock An efficient joint formulation for bayesian face verification.
\newblock {\em IEEE Transactions on Pattern Analysis and Machine Intelligence},
  2017.

\bibitem{ArcFace}
Jiankang Deng, Jia Guo, Niannan Xue, and Stefanos Zafeiriou.
\newblock Arcface: Additive angular margin loss for deep face recognition.
\newblock In {\em IEEE Conference on Computer Vision and Pattern Recognition},
  2019.

\bibitem{marginal_loss}
Jiankang Deng, Yuxiang Zhou, and Stefanos Zafeiriou.
\newblock Marginal loss for deep face recognition.
\newblock In {\em IEEE Conference on Computer Vision and Pattern Recognition
  Workshops}, 2017.

\bibitem{uniformface}
Yueqi Duan, Jiwen Lu, and Jie Zhou.
\newblock Uniformface: Learning deep equidistributed representation for face
  recognition.
\newblock In {\em IEEE Conference on Computer Vision and Pattern Recognition},
  2019.

\bibitem{msceleb1m}
Yandong Guo, Lei Zhang, Yuxiao Hu, Xiaodong He, and Jianfeng Gao.
\newblock Ms-celeb-1m: A dataset and benchmark for large-scale face
  recognition.
\newblock In Bastian Leibe, Jiri Matas, Nicu Sebe, and Max Welling, editors,
  {\em European Conference on Computer Vision}, 2016.

\bibitem{he2017mask}
Kaiming He, Georgia Gkioxari, Piotr Doll{\'a}r, and Ross Girshick.
\newblock Mask r-cnn.
\newblock In {\em IEEE International Conference on Computer Vision}, 2017.

\bibitem{resnet}
Kaiming He, Xiangyu Zhang, Shaoqing Ren, and Jian Sun.
\newblock Deep residual learning for image recognition.
\newblock In {\em IEEE Conference on Computer Vision and Pattern Recognition},
  2016.

\bibitem{LFW}
Gary~B. Huang, Manu Ramesh, Tamara Berg, and Erik Learned-Miller.
\newblock Labeled faces in the wild: A database for studying face recognition
  in unconstrained environments.
\newblock Technical report, University of Massachusetts, Amherst, 2007.

\bibitem{AFRN}
Bong-Nam Kang, Yonghyun Kim, Bongjin Jun, and Daijin Kim.
\newblock Attentional feature-pair relation networks for accurate face
  recognition.
\newblock In {\em IEEE International Conference on Computer Vision}, 2019.

\bibitem{PRN}
Bong-Nam Kang, Yonghyun Kim, and Daijin Kim.
\newblock Pairwise relational networks for face recognition.
\newblock In {\em European Conference on Computer Vision}, 2018.

\bibitem{kemelmacher2016megaface}
Ira Kemelmacher-Shlizerman, Steven~M Seitz, Daniel Miller, and Evan Brossard.
\newblock The megaface benchmark: 1 million faces for recognition at scale.
\newblock In {\em IEEE Conference on Computer Vision and Pattern Recognition},
  2016.

\bibitem{5459250}
N. {Kumar}, A.~C. {Berg}, P.~N. {Belhumeur}, and S.~K. {Nayar}.
\newblock Attribute and simile classifiers for face verification.
\newblock In {\em IEEE International Conference on Computer Vision}, 2009.

\bibitem{adaptiveface}
Hao Liu, Xiangyu Zhu, Zhen Lei, and Stan~Z Li.
\newblock Adaptiveface: Adaptive margin and sampling for face recognition.
\newblock In {\em IEEE Conference on Computer Vision and Pattern Recognition},
  2019.

\bibitem{SphereFace}
W. {Liu}, Y. {Wen}, Z. {Yu}, M. {Li}, B. {Raj}, and L. {Song}.
\newblock Sphereface: Deep hypersphere embedding for face recognition.
\newblock In {\em IEEE Conference on Computer Vision and Pattern Recognition},
  2017.

\bibitem{t_SNE}
Laurens van~der Maaten and Geoffrey Hinton.
\newblock Visualizing data using t-sne.
\newblock {\em Journal of machine learning research}, 2008.

\bibitem{IJB_C}
B. {Maze}, J. {Adams}, J.~A. {Duncan}, N. {Kalka}, T. {Miller}, C. {Otto},
  A.~K. {Jain}, W.~T. {Niggel}, J. {Anderson}, J. {Cheney}, and P. {Grother}.
\newblock Iarpa janus benchmark - c: Face dataset and protocol.
\newblock In {\em International Conference on Biometrics}, 2018.

\bibitem{age_db}
Stylianos Moschoglou, Athanasios Papaioannou, Christos Sagonas, Jiankang Deng,
  Irene Kotsia, and Stefanos Zafeiriou.
\newblock Agedb: the first manually collected, in-the-wild age database.
\newblock In {\em IEEE Conference on Computer Vision and Pattern Recognition
  Workshops}, 2017.

\bibitem{CSML}
Hieu~V. Nguyen and Li Bai.
\newblock Cosine similarity metric learning for face verification.
\newblock In Ron Kimmel, Reinhard Klette, and Akihiro Sugimoto, editors, {\em
  Asian Conference on Computer Vision}, 2011.

\bibitem{noroozi2016unsupervised}
Mehdi Noroozi and Paolo Favaro.
\newblock Unsupervised learning of visual representations by solving jigsaw
  puzzles.
\newblock In {\em European Conference on Computer Vision}, 2016.

\bibitem{vggface}
Omkar~M. Parkhi, Andrea Vedaldi, and Andrew Zisserman.
\newblock Deep face recognition.
\newblock In {\em British Machine Vision Conference}, 2015.

\bibitem{FaceNet}
F. {Schroff}, D. {Kalenichenko}, and J. {Philbin}.
\newblock Facenet: A unified embedding for face recognition and clustering.
\newblock In {\em IEEE Conference on Computer Vision and Pattern Recognition},
  2015.

\bibitem{CFP_FP}
S. {Sengupta}, J. {Chen}, C. {Castillo}, V.~M. {Patel}, R. {Chellappa}, and
  D.~W. {Jacobs}.
\newblock Frontal to profile face verification in the wild.
\newblock In {\em IEEE Winter Conference on Applications of Computer Vision},
  2016.

\bibitem{fisherface}
Karen Simonyan, Omkar Parkhi, Andrea Vedaldi, and Andrew Zisserman.
\newblock Fisher vector faces in the wild.
\newblock In {\em British Machine Vision Conference}, 2013.

\bibitem{contrastive_loss}
Yi Sun, Yuheng Chen, Xiaogang Wang, and Xiaoou Tang.
\newblock Deep learning face representation by joint
  identification-verification.
\newblock In {\em International Conference on Neural Information Processing
  Systems}, 2014.

\bibitem{deepface}
Y. {Taigman}, M. {Yang}, M. {Ranzato}, and L. {Wolf}.
\newblock Deepface: Closing the gap to human-level performance in face
  verification.
\newblock In {\em IEEE Conference on Computer Vision and Pattern Recognition},
  2014.

\bibitem{eigenface}
M.~A. {Turk} and A.~P. {Pentland}.
\newblock Face recognition using eigenfaces.
\newblock In {\em IEEE Conference on Computer Vision and Pattern Recognition},
  1991.

\bibitem{imdbface}
Fei Wang, Liren Chen, Cheng Li, Shiyao Huang, Yanjie Chen, Chen Qian, and
  Chen~Change Loy.
\newblock The devil of face recognition is in the noise.
\newblock In Vittorio Ferrari, Martial Hebert, Cristian Sminchisescu, and Yair
  Weiss, editors, {\em European Conference on Computer Vision}, 2018.

\bibitem{NormFace}
Feng Wang, Xiang Xiang, Jian Cheng, and Alan~Loddon Yuille.
\newblock Normface: L2 hypersphere embedding for face verification.
\newblock In {\em ACM International Conference on Multimedia}, 2017.

\bibitem{CosFace}
H. {Wang}, Y. {Wang}, Z. {Zhou}, X. {Ji}, D. {Gong}, J. {Zhou}, Z. {Li}, and W.
  {Liu}.
\newblock Cosface: Large margin cosine loss for deep face recognition.
\newblock In {\em IEEE Conference on Computer Vision and Pattern Recognition},
  2018.

\bibitem{centerloss}
Yandong Wen, Kaipeng Zhang, Zhifeng Li, and Yu Qiao.
\newblock A discriminative feature learning approach for deep face recognition.
\newblock In Bastian Leibe, Jiri Matas, Nicu Sebe, and Max Welling, editors,
  {\em European Conference on Computer Vision}, 2016.

\bibitem{IJB_B}
C. {Whitelam}, E. {Taborsky}, A. {Blanton}, B. {Maze}, J. {Adams}, T. {Miller},
  N. {Kalka}, A.~K. {Jain}, J.~A. {Duncan}, K. {Allen}, J. {Cheney}, and P.
  {Grother}.
\newblock Iarpa janus benchmark-b face dataset.
\newblock In {\em IEEE Conference on Computer Vision and Pattern Recognition
  Workshops}, 2017.

\bibitem{YTF}
L. {Wolf}, T. {Hassner}, and I. {Maoz}.
\newblock Face recognition in unconstrained videos with matched background
  similarity.
\newblock In {\em IEEE Conference on Computer Vision and Pattern Recognition},
  2011.

\bibitem{WHT:ECCVW08:DBMW}
L. Wolf, T. Hassner, and Y. Taigman.
\newblock Descriptor based methods in the wild.
\newblock In {\em European Conference on Computer Vision Workshops}, 2008.

\bibitem{ComparatorNet}
Weidi Xie, Li Shen, and Andrew Zisserman.
\newblock Pairwise relational networks for face recognition.
\newblock In {\em European Conference on Computer Vision}, 2018.

\bibitem{yang2016joint}
Jianwei Yang, Devi Parikh, and Dhruv Batra.
\newblock Joint unsupervised learning of deep representations and image
  clusters.
\newblock In {\em IEEE Conference on Computer Vision and Pattern Recognition},
  2016.

\bibitem{casia_webface}
Dong Yi, Zhen Lei, Shengcai Liao, and Stan~Z. Li.
\newblock Learning face representation from scratch.
\newblock {\em CoRR}, 2014.

\bibitem{QiYin:2011:AMF:2191740.2192084}
Qi Yin, Xiaoou Tang, and Jian Sun.
\newblock An associate-predict model for face recognition.
\newblock In {\em IEEE Conference on Computer Vision and Pattern Recognition},
  2011.

\bibitem{range_loss}
X. {Zhang}, Z. {Fang}, Y. {Wen}, Z. {Li}, and Y. {Qiao}.
\newblock Range loss for deep face recognition with long-tailed training data.
\newblock In {\em IEEE International Conference on Computer Vision}, 2017.

\bibitem{regularface}
Kai Zhao, Jingyi Xu, and Ming-Ming Cheng.
\newblock Regularface: Deep face recognition via exclusive regularization.
\newblock In {\em IEEE Conference on Computer Vision and Pattern Recognition},
  2019.

\bibitem{CPLFWTech}
T. Zheng and W. Deng.
\newblock Cross-pose lfw: A database for studying cross-pose face recognition
  in unconstrained environments.
\newblock Technical Report 18-01, Beijing University of Posts and
  Telecommunications, February 2018.

\bibitem{CALFW}
Tianyue Zheng, Weihong Deng, and Jiani Hu.
\newblock Cross-age {LFW:} {A} database for studying cross-age face recognition
  in unconstrained environments.
\newblock {\em CoRR}, abs/1708.08197, 2017.

\bibitem{ring_loss}
Y. {Zheng}, D.~K. {Pal}, and M. {Savvides}.
\newblock Ring loss: Convex feature normalization for face recognition.
\newblock In {\em IEEE Conference on Computer Vision and Pattern Recognition},
  2018.

\end{thebibliography}
}

\end{document}